\documentclass[conference]{IEEEtran}
\usepackage{amsmath,amssymb,amsfonts}
\usepackage{algorithmic}
\usepackage{graphicx}
\usepackage{natbib}
\usepackage{booktabs}
\usepackage{array}
\usepackage{url}
\usepackage{textcomp}
\usepackage{xurl}
\usepackage{pifont}
\usepackage{xcolor}
\usepackage{tcolorbox}
\usepackage{pdfpages}
\usepackage{tikz}
\usetikzlibrary{arrows.meta,positioning}
\tcbuselibrary{skins}

\definecolor{okgreen}{RGB}{0,130,60}
\definecolor{badred}{RGB}{190,30,40}
\definecolor{stepfill}{RGB}{243,246,249}
\definecolor{stepline}{RGB}{90,110,130}

\newtcolorbox{resbox}[1][]{colback=black!5,colframe=black!35,arc=0.8mm,
  boxrule=0.4pt,left=2.5pt,right=2.5pt,top=2pt,bottom=2pt,#1}
\newtcolorbox{stepbox}[1]{colback=stepfill,colframe=stepline,title={#1},
  fonttitle=\bfseries\footnotesize,arc=1.5mm,boxrule=0.6pt,
  left=3pt,right=3pt,top=2pt,bottom=2pt,toptitle=1pt,bottomtitle=1pt}
\newtcolorbox{infobox}[2][]{colback=#2!6,colframe=#2!50!black,
  fonttitle=\bfseries\footnotesize,arc=1.5mm,boxrule=0.6pt,
  left=3pt,right=3pt,top=2pt,bottom=2pt,toptitle=1pt,bottomtitle=1pt,#1}

\newcommand{\stepnum}[1]{\tikz[baseline=(c.base)]{\node[circle,fill=white,
  draw=stepline,text=stepline,inner sep=1.1pt,font=\tiny\bfseries](c){#1};}}
\newcommand{\goldbadge}[2]{\colorbox{#1}{\textcolor{white}%
  {\bfseries\tiny~Gold: #2~}}}
\newcommand{\zone}[2]{\begin{minipage}[t][#1][t]{\linewidth}#2\end{minipage}}
\newcommand{\point}[1]{\vskip 0.05in \noindent #1 \par}

\IEEEoverridecommandlockouts
\def\BibTeX{{\rm B\kern-.05em{\sc i\kern-.025em b}\kern-.08em
    T\kern-.1667em\lower.7ex\hbox{E}\kern-.125emX}}
    
\begin{document}

\title{Turkish Multimodal Video Fact-Checking: A New Dataset and a Comparison of Zero-Shot Agentic and Fine-Tuned Systems\\
}

\author{\IEEEauthorblockN{Azra Açıl}
\IEEEauthorblockA{
\textit{MSc Artificial Intelligence, QMUL} \\
a.acil@se25.qmul.ac.uk \\
Project Supervisor: Professor Shalom Lappin \\
s.lappin@qmul.ac.uk}}

\maketitle

\begin{abstract}
Video-based misinformation spreads rapidly on social media, but almost all existing fact-checking systems focus exclusively on English text. Tools for low-resource languages such as Turkish are severely limited. In this work, we present TeViD (Turkish Video Fact-checking Dataset), to the best of our knowledge the first multimodal video fact-checking dataset for Turkish. We compiled the dataset from articles published by Teyit.org, an IFCN-certified organisation, and enriched it with translated Snopes articles and synthetic examples to reduce class imbalance. Building on TeViD, we develop two verification pipelines that share the same open-source base model. First, in 3MFact-TR (Pipeline 1), we adapt the multi-role 3MFact framework to Turkish with the Qwen2.5-7B large language model and the Qwen2-VL-7B vision-language model, and add three modules of our own: a semantic inconsistency checker, a 5W1H question manager, and a bilingual evidence retriever that excludes fact-checking sites to prevent leakage. Second, as our Fine-Tuned Model (Pipeline 2), we fine-tune a single Qwen2.5-7B with Low-Rank Adaptation (LoRA) to produce closed-book verdicts from the claim and the video's textual description. We report accuracy, macro F1, True class F1, and True class recall for both systems, together with ablation studies. 
\end{abstract}

\begin{IEEEkeywords}
fact-checking, multimodal verification, misinformation detection, Turkish
NLP, low-resource languages, large language models
\end{IEEEkeywords}

\section{Introduction}
Misinformation has long been a problem, but it has become increasingly important with the spread of social media and AI-generated content. False stories no longer travel only as text; they spread far faster and more convincingly through video. Verifying a video is harder than verifying text, because one must judge at the same time what the footage shows, what is said in the audio, and if the claim is consistent with both. Professional fact checkers do this analysis by hand, but manual verification cannot scale against the volume of content produced today. A critical compounding issue is that almost all automated systems are English and text-based. Video-based verification remains understudied, and tools for low-resource languages like Turkish are nearly nonexistent.

Although there is a lot of research on multimodal misinformation detection, most of it focuses on either English or low-resource languages without video, not both. FACTIFY3M~\citep{chakraborty2023factify3m} provides explainability, but it is limited to text and static images. Fact-R1~\citep{zhang2025factr1} brings deep reasoning to video, yet relies on a large-scale English dataset. For Turkish the picture is even narrower: FCTR~\citep{cekinel2024fctr} is a valuable resource, but it only contains textual claims. The system closest to our scenario, 3MFact~\citep{niu2025three}, uses a multi-role architecture that analyses a video by dividing labour across different models. However, it also focuses only on English and depends on proprietary, paid application programming interface (API) models, which makes it expensive to use and unsuitable for Turkish content out of the box.

As a result, no existing approach addresses video, multilinguality, and evidence-based verification together, and Turkish has no labelled multimodal video fact-checking dataset suitable for evaluation. In this work we aim to fill this gap by adapting the 3MFact framework to Turkish using open-source models and building the first Turkish multimodal video fact-checking dataset. To achieve this goal, this study examines three fundamental questions. First, is it possible to create a multimodal video fact-checking dataset for a low-resource language? Second, can the multi-role 3MFact architecture be adapted to Turkish using only open models? Finally, how do the two pipelines compare, and which system achieves better performance on this task?

To answer these questions, our study makes three primary contributions. First, we introduce TeViD (Turkish Video Fact-checking Dataset), which, to the best of our knowledge, is the first multimodal video fact-checking dataset for Turkish. We generate the dataset from articles published by Teyit.org~\citep{teyit},
certified by the International Fact-Checking Network (IFCN;~\citealp{ifcn}), and enrich it with Snopes~\citep{snopes} articles and synthetic examples to reduce class imbalance. Second, we develop Pipeline 1 (3MFact-TR), a zero-shot agentic system that adapts the multi-role 3MFact framework to Turkish using the Qwen2.5-7B Large Language Model (LLM)~\citep{yang2024qwen25} and the Qwen2-VL-7B Vision-Language Model (VLM)~\citep{wang2024qwen2vl}. To this pipeline, we add three modules of our own: a Semantic Inconsistency Checker (SIC) to detect visual mismatches, a 5W1H question manager to structure queries, and a bilingual evidence retriever that strictly excludes fact-checking sites to prevent test data leakage. Finally, we introduce the Fine-Tuned Model (Pipeline 2), which is the same Qwen2.5-7B fine-tuned with Low-Rank Adaptation (LoRA)~\citep{hu2022lora} to act as a closed-book system that decides from its own knowledge without external evidence. Since both pipelines use the exact same base model, any difference in their performance comes directly from the system architecture. We evaluate both systems on the TeViD test set and report accuracy, macro F1, True class F1, and True class recall, together with ablation studies that isolate the contribution of each module. We then discuss our findings qualitatively in relation to the published performance of the original English system.

\section{Related Work}

\subsection{Multimodal and Video-Based Misinformation Detection}
Misinformation detection has long focused on text, with multimodal approaches emerging more recently. SAFE~\citep{zhou2020safe} captures inconsistencies by measuring the similarity between the text and the image of a news article, but it requires training and is limited to static images. QACheck~\citep{pan2023qacheck} breaks verification into a question guided multi-step process, following a logic similar to our own question manager, yet it operates on text alone. FACTIFY3M~\citep{chakraborty2023factify3m} is a large dataset that offers explainability through 5W question answering, but it too remains limited to text and static images and does not cover video. Its 5W scheme is the starting point for our own question manager, which we recast as an inference-time question generator and extend with How (Section~IV-B). On the video side, Fact-R1~\citep{zhang2025factr1} brings deep reasoning to video, but it requires heavy reinforcement learning on a large-scale English dataset and does not adapt easily to an open, non-proprietary setup. What these works have in common is that they do not address video, multilinguality, and evidence-based verification together. Our work targets precisely the intersection of these three.

\subsection{Turkish Verification Resources}
Several resources have been introduced for Turkish misinformation research in recent years, but none of them cover video-based multimodal verification. FCTR~\citep{cekinel2024fctr} is our closest precedent, testing cross-lingual transfer against low-resource fine-tuning using three Turkish fact-checking organisations. Unlike our system, it relies on text-only claims and lacks live evidence retrieval. MiDe22~\citep{toraman2022mide22} is a misinformation dataset of Turkish and English tweets labelled with engagement data; although the tweets sometimes contain images, the focus is not on modelling the audiovisual stream of a video. MSVD-Turkish~\citep{citamak2021msvdturkish} provides video-text data for Turkish, but its task is video captioning, describing the content of a video, rather than fact-checking. M4FC~\citep{geng2025m4fc} is a large, real-world multimodal verification dataset spanning ten languages including Turkish, but its content consists of static images paired with claims and does not include a video dimension. In short, the picture for Turkish is narrow and video-based fact-checking has been left unaddressed. To the best of our knowledge, we present the first dataset to fill this gap.

\subsection{Reference Architecture: 3MFact}
The architecture our work builds on, 3MFact~\citep{niu2025three}, is a multi-role, zero-shot verification framework that analyses a video through a division of labour among several models. It consists of five components: a Video Descriptor that converts the video into text, a Claim Verifier that checks whether the available information is sufficient for a decision, a Question Manager that generates questions for the missing information, an Information Retriever that answers these questions through web search and selects evidence, and a Reasoner that combines all the evidence to produce the final verdict, the rationale, and the supporting evidence. The original system uses GPT-4o-mini as its language model and MiniCPM-V 2.6 as its vision-language model, and is designed only for English. We build on their public implementation. We replace the proprietary models with open-source ones, rewrite the prompts and the configuration for Turkish, substantially modify the orchestration and the reasoning stage, and add three
new modules. We do not re-run the original English system, and cite its
published figures only as a reference point~\citep{niu2025code}.

\section{Dataset}
\label{sec:dataset}
 
We constructed a new dataset of 2,097
fact-checked video claims compiled from three different sources. Each claim is paired with its target video, video metadata, a
transcript, an automatic visual description, evidence references, and a
reference rationale distilled from the fact-checker's published verdict. We
split the data into 1,738 training and 359 test examples (124 True, 235
False); Table~\ref{tab:dataset} summarises the overall composition.
 
\begin{table}[t]
\centering
\caption{\normalfont Dataset composition by source and label.}
\label{tab:dataset}
\begin{tabular}{lrrr}
\toprule
\textbf{Source} & \textbf{True} & \textbf{False} & \textbf{Total} \\
\midrule
Teyit.org (native)                 & 124 & 1,027 & 1,151 \\
Snopes (translated)                & 407 & 324   & 731   \\
Ensonhaber (synthetic) & 215 & 0     & 215   \\
\midrule
\textbf{Total} & \textbf{746} & \textbf{1,351} & \textbf{2,097} \\
\bottomrule
\end{tabular}
\end{table}

\subsection{Sources and Ethics}
Our primary source is Teyit.org, the largest fact-checking organisation in Turkey and a signatory of the IFCN Code of Principles. Professional fact-checkers assign labels, and before collecting any data we obtained the necessary permission for academic use from the organisation's editor. Our second source is the TRUE dataset compiled by the original 3MFact work~\citep{niu2025three} from Snopes.com, which is also an IFCN signatory. We selected articles from this dataset, translated them into Turkish with machine translation, and included them in our dataset. Our third source is a set of synthetic ``True'' articles created from real footage and recordings published on Ensonhaber.com~\citep{ensonhaber}, covering accidents, rescue operations, and natural events. These articles are used only in the training split. During data collection, we filtered out sensitive content such as violent material.

\subsection{Teyit.org and Pre-processing}

We collected the Teyit.org articles by scraping the site, obtaining 2,473
articles that contain at least one video. From each article we took the
fact-checker's verdict, a single target video, and the cited evidence
references. We mapped these fields onto the English TRUE dataset's schema to
ensure that they align with our other sources.

The scraped text included a lot of irrelevant information. We removed player
artefacts, boilerplate, and author names from the article body. We also
normalised the rationales into a consistent form for later use as reference
text.

We then filtered the dataset to items usable by a video pipeline. Since the
pipeline works in Turkish and English, we removed videos whose titles are in
a non-Latin script. We eliminated sensitive content and duplicate videos. We
kept only articles whose video metadata could be recovered with
\texttt{yt-dlp}, an open-source video retrieval tool~\citep{ytdlp2024}. The articles that pass these stages form the Teyit.org row of
Table~\ref{tab:dataset}.

Only 10.8\% of the native Teyit articles were labelled as True (124 of 1,151), because the organisation publishes far more debunks than confirmations. An imbalance of
this size would make accuracy misleading: a model that labels every item
``False'' without ever catching a single ``True'' claim would still reach
roughly 89\% accuracy on this collection. Therefore, we adopt macro F1 as our
primary metric and address the imbalance using the augmentation strategies
described in the next section.

\subsection{Cross-Lingual Augmentation with Snopes}

To strengthen the True class, we used the verified True claims in the TRUE dataset (Snopes). However, many of these articles concern local events known only in the United States, which would not appear in a Turkish-language search. Therefore, we filtered them in two stages. In the first stage, we applied a coarse filter using more than 150 US-specific keywords, such as state names, local politicians, and institutions. In the second stage, we classified each remaining claim as global or US-local with a Qwen2.5-7B model. We only asked the model whether the claim's topic was universal, not whether it was true.

We translated the remaining True articles into Turkish with NLLB-200-1.3B, a multilingual machine-translation model~\citep{nllbteam2022nllb}. We deliberately did the translation using a different model because the pipeline uses Qwen2.5-7B. This ensures that during inference, the model does not evaluate text it generated itself.

These translations created a new problem. Every translated Snopes article was True, while almost every Teyit article was False. Therefore, a model could learn a shortcut based only on the source rather than the content of the claim: True if translated, False if from Teyit. To break this correlation, we translated a comparable number of False articles using the same process. This balanced the Snopes portion between True and False (Table~\ref{tab:dataset}).

\subsection{Synthetic Turkish True Examples}

The translated examples corrected the balance, but they were not native
Turkish. We used news videos from Ensonhaber.com to include True examples that are genuinely Turkish and contain real event footage. We scanned the site's news videos and manually reviewed roughly 1,200 candidates, selecting 215 clips that show real events such as accidents, rescues, and natural events; we eliminated interview and studio content.

To prevent superficial shortcuts, we processed each clip individually. This
included trimming the channel bumpers, cropping the logo and bottom banners,
and saving the clips in a 50/30/20 ratio (original/480p/360p). This prevented
a class cue such as ``clean, high-resolution footage means True'' from
forming.

For each clip, we generated a single-sentence viral claim that corresponds to
the news headline and summary. For the same reason as in the translation step, we used Gemma 2 9B, an open LLM from a different model family~\citep{gemmateam2024gemma2}, as the generator. We anchored the prompts with a small number of examples sampled from real Teyit claims. The claims were required to be in the past tense, without exaggeration or any indication of a verdict, with any news channel names masked.

We generated the video descriptions with the same module used for the rest of
the dataset, so that the synthetic descriptions share the same distribution as the real ones. We used these articles in the training split only, and none of them enter the evaluation.

\subsection{Enrichment, Splitting, and Leakage Controls}

Each article contains a transcript, video description, evidence references,
and a gold rationale. The transcripts were created using faster-whisper
large-v3~\citep{fasterwhisper2023}, an optimised reimplementation of Whisper~\citep{radford2023whisper}, under voice activity detection (VAD)~\citep{silerovad2024}, with zero temperature and no
context conditioning. We explicitly marked silent videos as ``no speech''.
For each article video, we generated a single-paragraph Turkish description
with a single Qwen2-VL call over eight equally spaced frames. We generated the reference rationales, called LLM-Summary Rationales (LSR) in the TRUE schema, with Qwen2.5-7B. We detected corrupted LSR outputs that drifted into another language or simply restated the claim using pattern matching, then rectified them using a validation filter. 

The split is not random. We wanted to measure what the pipelines do on native True claims. So we put all 124 of them in the test portion. We then added
native False articles in proportion, which gives the 359 article test set. The
remaining 792 native articles are all False. They stay in training with the
translated and the synthetic ones. This is why Pipeline~2 trains inside the
evaluation pool rather than on the outer split (Section~IV-C).

We implement strict controls against four potential sources of data leakage. First, to prevent training and test leakage in TeViD, we deduplicated claims and videos between splits, masked organisation names while keeping the Turkish verb \emph{teyit etmek} (``to confirm''), and locked the frozen test set with a SHA-256 digest. Second, to address potential pretraining contamination, we implemented perplexity and context-free verdict tests. Teyit claims had higher perplexity under Qwen2.5-7B than unseen control claims (106.9 vs 45.5), which is the opposite of memorisation behaviour, and the model failed to exceed the majority class baseline without context. Third, to prevent retrieval leakage during verification, Pipeline 1 strictly blocks fact-checking domains and URL traces (Section~\ref{sec:pipeline}). A review of the final results showed that no fact-checking domains were selected or downloaded during the evaluation. Finally, to prevent self-evaluation bias, we isolated the augmentation models by using NLLB-200 and Gemma~2, ensuring the Qwen-based pipeline never evaluates text it generated itself. 

Following all data processing and splitting, one article was excluded from the final test set due to a video download failure, leaving exactly 358 articles for the evaluation pool discussed in Section~\ref{sec:results}.

\section{System}
\label{sec:system}
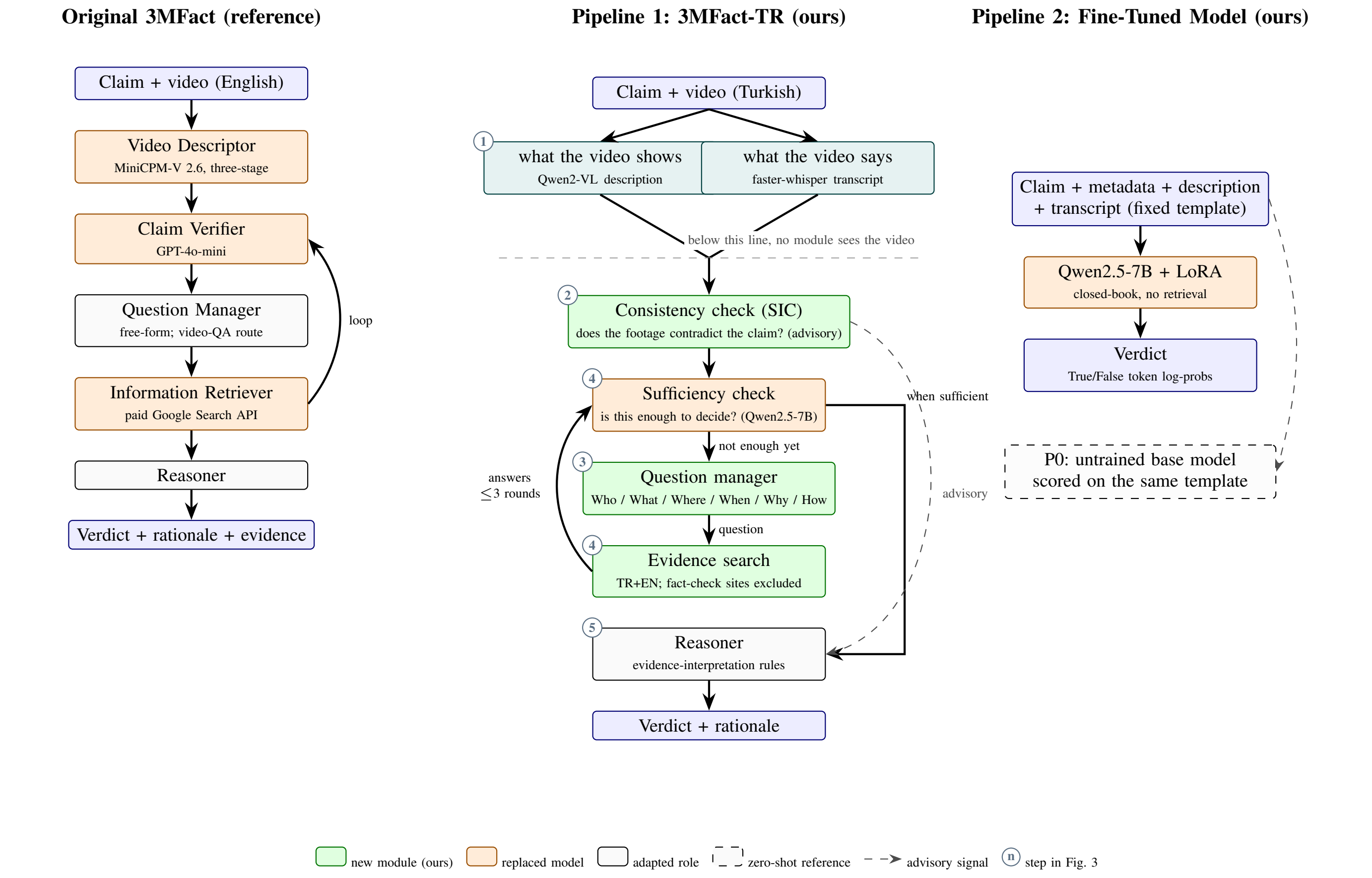
\begin{figure*}[t]
\centering
\begin{tikzpicture}[
  box/.style={draw, rounded corners=1.5pt, align=center, font=\scriptsize,
              inner sep=3pt, minimum width=3.1cm, fill=gray!4},
  new/.style={box, fill=green!12, draw=green!45!black},
  rep/.style={box, fill=orange!15, draw=orange!60!black},
  io/.style={box, fill=blue!7, draw=blue!45!black},
  off/.style={box, fill=teal!10, draw=teal!50!black, text width=2.5cm},
  badge/.style={circle, draw=stepline, fill=white, text=stepline,
                inner sep=1.1pt, font=\tiny\bfseries},
  q/.style={font=\tiny},
  arr/.style={-Stealth, thick},
  adv/.style={-Stealth, dashed, gray!60!black},
  node distance=4mm
]

% ---------- Sutun 1: Orijinal 3MFact (rol adlari orijinaldeki gibi) ----------
\node[font=\footnotesize\bfseries] (t3) at (0,0) {Original 3MFact (reference)};
\node[io, below=of t3] (a1) {Claim + video (English)};
\node[rep, below=of a1] (a2) {Video Descriptor\\ {\tiny MiniCPM-V 2.6, three-stage}};
\node[rep, below=of a2] (a3) {Claim Verifier\\ {\tiny GPT-4o-mini}};
\node[box, below=of a3] (a4) {Question Manager\\ {\tiny free-form; video-QA route}};
\node[rep, below=of a4] (a5) {Information Retriever\\ {\tiny paid Google Search API}};
\node[box, below=of a5] (a6) {Reasoner};
\node[io, below=of a6] (a7) {Verdict + rationale + evidence};
\draw[arr] (a1) -- (a2); \draw[arr] (a2) -- (a3); \draw[arr] (a3) -- (a4);
\draw[arr] (a4) -- (a5); \draw[arr] (a5) -- (a6); \draw[arr] (a6) -- (a7);
\draw[arr] (a5.east) to[bend right=40] node[right,q]{loop} (a3.east);

% ---------- Sutun 2: Pipeline 1 (fonksiyon adlari, tablo diliyle) ----------
\node[font=\footnotesize\bfseries] (t1) at (6.9,0) {Pipeline 1: 3MFact-TR (ours)};
\node[io] (b1) at (6.9,-1.0) {Claim + video (Turkish)};
\node[off] (b2a) at (5.45,-2.0) {what the video shows\\ {\tiny Qwen2-VL description}};
\node[off] (b2b) at (8.35,-2.0) {what the video says\\ {\tiny faster-whisper transcript}};
\draw[arr] (b1.south) -- (b2a.north);
\draw[arr] (b1.south) -- (b2b.north);
\coordinate (m) at (6.9,-3.2);
\draw[thick] (b2a.south) -- (m);
\draw[thick] (b2b.south) -- (m);
% metinlesme ayraci — yazi cizginin USTUNDE, beyaz dolgulu (cakismasin)
\draw[dashed, gray!70] (4.1,-3.2) -- (9.7,-3.2);
\node[q, gray!60!black, anchor=south east, fill=white, inner sep=1.5pt]
  at (9.7,-3.10) {below this line, no module sees the video};
\node[new] (b3) at (6.9,-4.05) {Consistency check (SIC)\\
  {\tiny does the footage contradict the claim? (advisory)}};
\node[rep, below=of b3] (b4) {Sufficiency check\\
  {\tiny is this enough to decide? (Qwen2.5-7B)}};
\node[new, below=of b4] (b5) {Question manager\\
  {\tiny Who / What / Where / When / Why / How}};
\node[new, below=of b5] (b6) {Evidence search\\
  {\tiny TR+EN; fact-check sites excluded}};
\node[box, below=of b6] (b7) {Reasoner\\
  {\tiny evidence-interpretation rules}};
\node[io, below=of b7] (b8) {Verdict + rationale};
\draw[arr] (m) -- (b3.north);
\draw[arr] (b3) -- (b4);
\draw[arr] (b4) -- node[right,q]{not enough yet} (b5);
\draw[arr] (b5) -- node[right,q]{question} (b6);
\draw[arr] (b4.east) -- ++(1.05,0)
  node[right,q,pos=0.9,yshift=1.2mm]{when sufficient}
  |- (b7.east);
\draw[arr] (b7) -- (b8);
\draw[arr] (b6.west) to[bend left=45]
  node[left,q,align=center,xshift=-1mm]{answers\\ $\leq$3 rounds} (b4.west);
\draw[adv] (b3.east) to[bend left=70]
  node[right,q,xshift=0.5mm]{advisory} (b7.east);
% rozetler: Fig. 2 adimlari (asagidan yukari akis sirasi)
% Rozetler: Fig. 2'deki adim numaralariyla birebir.
% 4 iki kutuda: Fig. 2'nin 4. adimi yeterlilik ile aramayi birlikte anlatiyor.
\node[badge] at (b2a.north west) {1};
\node[badge] at (b3.north west)  {2};
\node[badge] at (b4.north west)  {4};
\node[badge] at (b5.north west)  {3};
\node[badge] at (b6.north west)  {4};
\node[badge] at (b7.north west)  {5};

% ---------- Sutun 3: Pipeline 2 + P0 ----------
\node[font=\footnotesize\bfseries] (t2) at (12.65,0) {Pipeline 2: Fine-Tuned Model (ours)};
\node[io, below=18mm of t2] (c1) {Claim + metadata + description\\ + transcript (fixed template)};
\node[rep, below=of c1] (c2) {Qwen2.5-7B + LoRA\\ {\tiny closed-book, no retrieval}};
\node[io, below=of c2] (c3) {Verdict\\ {\tiny True/False token log-probs}};
\draw[arr] (c1) -- (c2); \draw[arr] (c2) -- (c3);
\node[box, dashed, below=7mm of c3, fill=gray!2, text width=3.4cm] (c4)
  {P0: untrained base model\\ scored on the same template};
\draw[adv] (c1.east) to[bend left=17] (c4.east);

% ---------- Lejant ----------
\node[align=center, font=\tiny] at (6.9,-11.2) {
  \tikz\node[new, minimum width=4mm, minimum height=2.5mm]{}; new module (ours)\quad
  \tikz\node[rep, minimum width=4mm, minimum height=2.5mm]{}; replaced model\quad
  \tikz\node[box, minimum width=4mm, minimum height=2.5mm]{}; adapted role\quad
  \tikz\node[box, dashed, minimum width=4mm, minimum height=2.5mm]{}; zero-shot reference\quad
  \raisebox{1pt}{\tikz\draw[adv] (0,0) -- (5mm,0);} advisory signal\quad
  \tikz\node[badge]{n}; step in Fig.~\ref{fig:worked}
};
\end{tikzpicture}
\caption{\normalfont The architectures compared in this work. Pipeline 1 adapts the
original 3MFact by replacing its proprietary models with open-source
alternatives and adding three new modules; Pipeline 2 shares the same base
model but decides closed-book after LoRA fine-tuning; the original English
3MFact is used unmodified as a reference point. The video is turned into text
once, at the top: below the dashed line no module sees the video again, and
all reasoning operates on the two texts produced there. Circled numbers link
each box to the corresponding step of the worked example in
Fig.~\ref{fig:worked}, where the sufficiency check and the evidence search
are narrated together as step~4. The dashed box is P0,
an untrained copy of the base model scored on the same template as
Pipeline 2; it is a reference point, not a system we propose.}
\label{fig:arch}
\end{figure*}

We build two systems on TeViD that share the same open-source base model,
the large language model Qwen2.5-7B-Instruct: a zero-shot agentic pipeline that adapts 3MFact to Turkish, and a fine-tuned closed-book model that isolates the contribution of the agentic design. We also score an untrained copy of the same base model on the same input. We call it P0 and use it only as a zero-shot reference point. Figure~\ref{fig:arch} compares the two systems with the original architecture. Table~\ref{tab:components} lists what each component receives and produces.

\subsection{Shared Offline Pre-processing}
\label{sec:preproc}
Both systems read the video in the same way. The input to both systems is a
claim, a target video, and the video's metadata. Each video is turned into
text once, before verification begins.

\point{\textbf{What The Video Shows}} The vision-language model
Qwen2-VL-7B-Instruct views eight frames spread evenly across the clip and
generates a single Turkish paragraph describing what is shown. The prompt
requires that any on-screen text, such as banners, captions, and signs, be
quoted verbatim. We also specifically forbid identity guesses, because VLM face recognition is unreliable, and prohibit speculation about details that are not clearly visible. Step~1 of Figure~\ref{fig:worked} (Appendix~\ref{app:worked}) shows an example of this: the description captures the wave patterns and the tourists, but it cannot independently determine the location of the sandbank.

\point{\textbf{What The Video Says}} We transcribe the audio using the speech recognition model faster-whisper. Silent videos are marked explicitly
rather than left empty.

\point{\textbf{What The Rest Of The System Sees}} These steps produce two short texts. From this point on, no component looks at the video again. Checking
consistency, asking questions, searching the web, and reaching a verdict all
operate on these two texts, together with the claim and the video's metadata.
The visual and spoken streams are therefore never merged into a single
representation. They remain two separate texts that the language model reads
side by side.  

% ###########################################################################
% BLOK A  —  FIGURE 1 (yeni)
% ###########################################################################

% ###########################################################################
% BLOK B  —  BILESEN TABLOSU
% ###########################################################################

\begin{table}[t]
\caption{\normalfont What each component receives and produces. The first two
rows run once, offline; the consistency check runs before the loop and the
next three rows form it; the last two rows are the closed-book systems, which
skip the loop entirely.}
\label{tab:components}
\centering
\footnotesize
\begin{tabular}{@{}>{\raggedright\arraybackslash}p{2.05cm}%
>{\raggedright\arraybackslash}p{2.3cm}%
>{\raggedright\arraybackslash}p{3.45cm}@{}}
\toprule
\textbf{Component} & \textbf{Receives} & \textbf{Produces}\\
\midrule
\multicolumn{3}{@{}l}{\emph{Offline, once per video}}\\
Video description & eight frames & one paragraph: what is on screen\\
Transcript & the audio & what is said\\
\midrule
\multicolumn{3}{@{}l}{\emph{Before the loop}}\\
Consistency check (SIC) & claim, description, headline & a note: does the
footage contradict the claim?\\
\midrule
\multicolumn{3}{@{}l}{\emph{Pipeline 1: the verification loop}}\\
Sufficiency check & everything gathered so far & enough / not enough, and
how sure\\
Question manager & the claim, what is still missing & one question, tied to
Who, What, Where, When, Why, or How\\
Evidence search & the question & up to three web pages; fact-checking sites
excluded\\
Reasoner & all of the above & True or False, a misinformation type, and a rationale\\
\midrule
\multicolumn{3}{@{}l}{\emph{Closed book, no loop}}\\
Fine-tuned model (P2) & the same texts in a fixed template & True or False\\
Untrained base (P0) & the same template & True or False\\
\bottomrule
\end{tabular}
\end{table}

\subsection{Pipeline 1: The 3MFact-TR Pipeline}
\label{sec:pipeline}
 
Pipeline 1 keeps the multi-role architecture of 3MFact but replaces the proprietary models with the open-source ones named above. All behaviour is
controlled by prompts, and no role is trained. The output is a binary verdict with a rationale that is prompted to cite the retrieved evidence. The verification loop runs on the two offline texts until the verifier judges the evidence sufficient, up to a maximum of three rounds. We use greedy decoding and on-disk caches to ensure that the same input produces the same verdict in every run. Figure~\ref{fig:worked} traces both pipelines on TeViD test claims. The left-hand column follows a claim in which authentic footage is attached to a false location, and we return to it at each stage below.

\point{\textbf{Consistency Check (SIC)}}
Before the loop
begins, this module compares the video description against the claim and the video headline. It produces one of four outcomes: contradicts, not visible, unintelligible, or consistent. SIC is an advisory, not a gate. It cannot decide an article on its own, and its output is passed only to the final reasoning stage as an additional input. To prevent the module from reading the absence of evidence as a contradiction, we ask it first whether the claim and the footage could both be true at the same time. Only mutually exclusive scenarios are marked ``contradicts''. Otherwise the verdict is marked as ``not visible''. This distinction matters because descriptions are short and selective, and a detail that is not visible in the video does not make the claim false. The full prompt is given in Appendix~\ref{app:sicprompt}. In step~2 of Figure~\ref{fig:worked}, SIC reports \textsc{consistent} (0.8). The footage matches the claim visually, and a location lie is invisible to a purely visual check, so the decision is left to evidence.

\point{\textbf{Question Manager}} Unlike the original framework's free-form question generation, each question is committed to one dimension of the journalistic 5W1H scheme. This adapts the 5W question-answering idea of FACTIFY3M~\citep{chakraborty2023factify3m} into an inference-time question generator and extends it with How. We require the model to output the dimension before the question, and we track covered dimensions to prevent repetition. Questions must relate to the external world. A question is accepted only if a web search could answer it without watching the video, and video-directed questions (``what does the video show?'') are rejected. Greedy decoding would produce the identical output on a blind retry, so rejected questions are shown verbatim to the next attempt and the model is directed towards the remaining dimensions. In step~3 of Figure~\ref{fig:worked}, the primary question takes the Where dimension: whether the sandbank is located in the claimed Red Sea.

\point{\textbf{Sufficiency Check and Evidence Search}} These two stages
alternate. The verifier runs first. It determines whether the information gathered so far suffices to decide the claim, not whether the claim is true. It weighs the claim, the video description, the metadata, and the accepted question-answer pairs through chain-of-thought prompting, and produces a sufficiency judgment with a confidence value. The loop ends when the judgment is positive and the confidence exceeds 0.93. Otherwise control passes to the question manager, and the cycle repeats up to three times. For each accepted question the model produces exactly two search queries, one Turkish to reach local sources and one English to reach international coverage. A claim-derived anchor query is added in every round to prevent query drift. Fact-checking
domains and URL traces are blocked, so the system cannot rely on Teyit.org's
analysis directly. Before any page is downloaded, candidate results are re-ranked against the claim and the question with a cross-encoder (bge-reranker-v2-m3)~\citep{chen2024bge}. Downloaded contents are scored on a 0--10 scale for site quality, recency and
relevance, combined as a weighted sum with weights 0.25, 0.25 and 0.50, and the
three highest-scoring pieces of evidence are kept. Recency decreases with the
page's publication date, and evidence with no recoverable date scores zero. Each question is turned into an answer that cites this evidence, and a validator eliminates
unsuitable pairs. The original system had a secondary route that sent some
questions to the video instead of the web. We disabled it, because the
question manager already directs questions to the external world, and the
original ablation limits its contribution to approximately 3.5
points~\citep{niu2025three}. This configuration is labelled ``w/o QALMM''. In step~4 of Figure~\ref{fig:worked}, the recovered postings situate the
sandbank on Dongshan Island, Fujian, China (95\%), and the loop closes after
two follow-up questions.

\point{\textbf{Final Reasoning}} The Reasoner combines the claim, the video texts, the gathered question-answer pairs, and the SIC note into a binary verdict, a misinformation type, and a rationale. Its prompt contains specific evidence interpretation rules: the absence of evidence alone is not sufficient for ``False'', ``False'' is not a safe default, and verified external evidence supporting the claim outweighs the lack of visual confirmation. When an output cannot be parsed, the system does not generate a verdict. The article is returned with an explicit failure record and retried. In step~5 of Figure~\ref{fig:worked}, the Reasoner separates silence from
contradiction. The footage does not actively contradict the claim, but the
evidence identifies the true location, and the verdict is \textsc{False}
(false video description).

\subsection{Pipeline 2: The Fine-Tuned Model}
\label{sec:baseline}

The Fine-Tuned Model, shown in the right-hand column of Figure~\ref{fig:arch}, fine-tunes the same Qwen2.5-7B base model with LoRA into a closed-book verdict model. A closed-book system must decide the claim using only the provided input texts and its own parameters, without access to an external search engine or fact-checking database. The input is a fixed-slot Turkish template. It contains the claim, the video
title and date, the offline video description, and the transcript. Missing
fields carry an explicit placeholder rather than being omitted, and source
names are masked as in the rest of the dataset. The target is a single
verdict token, and the loss is computed only on the answer tokens. 

We apply LoRA with rank 16 (alpha 32, dropout 0.05) to all attention and MLP
projection matrices. We train in bf16 with a learning rate of $10^{-4}$
under cosine decay for a fixed budget of 120 steps. The recipe was fixed in
advance, with no hyperparameter tuning and no early stopping. At inference
the model generates no free text. The verdict is read by comparing the
log-probabilities of the True and False tokens at the first answer position,
which makes the decision deterministic and parse-free. We report the verdict
at $\tau=0$, that is, whichever token has the higher log-probability.

Section~\ref{sec:dataset} explains the split. Every native True article is in the evaluation pool, and there is no native True example outside it to train on. This is why Pipeline~2 trains inside the pool rather than on the outer split. The model is supervised, so it needs labelled examples that the evaluation does not reuse. We therefore apply five-fold stratified cross-validation inside the frozen native Teyit pool. Each fold trains on the remaining 286 articles and predicts the fold that was held out, and we pool the out-of-fold predictions. Every reported prediction therefore comes from a model that never saw that article during training. A tenth of each training fold is further reserved as a calibration slice and is never trained on. Four training conditions vary what else is added to the native articles and how long training runs: P2-A uses them alone, P2-B1 adds the 731 translated Snopes articles, and P2-B adds all remaining training data. A fourth condition, P2-A34, is P2-A trained for 34 steps instead of 120, which matches P2-B1's epoch count (Section~\ref{sec:results}).

The comparison between the two systems is inherently asymmetric. Pipeline 1 is a zero-shot framework that has never seen a labelled Turkish example. In contrast, the Fine-Tuned Model learns from native Turkish articles from the very same pool, even though it never trains on the exact article it evaluates. To measure how much this training helps, we use P0, the untrained base model, scored on the identical template.

% =====================================================================
% RESULTS - TEK SAYFALIK YOGUN SURUM (P1 + P2 birlikte)
% Sonradan genisletilecek cekirdek. Tek tablo, dort paragraf.
% =====================================================================

\section{Results}
\label{sec:results}

\subsection{Evaluation Protocol}
We evaluate all systems on the same frozen pool of 358 native Teyit articles
(124 True, 234 False; Section~\ref{sec:dataset}). Pipeline 1 runs on the full
359 article test split and returns one output that cannot be parsed, leaving 358 valid verdicts. The pool and Pipeline 1's valid set share 357 articles, and we run every paired test on those 357. We use macro-F1 as our primary metric. Accuracy is misleading on this pool, because a model that never predicts True still reaches 0.654. We also report
accuracy, F1 on the True class, and recall of the True class.

Every system sees the same articles, so we compare them with the exact
McNemar test. $b$ represents the number of articles where the first system is correct but the second is incorrect, and $c$ represents the opposite. All $p$ values are two-sided. We report a bootstrap 95\% confidence interval for macro-F1. We report all Pipeline 2 numbers as pooled out-of-fold predictions at $\tau=0$, so that no condition benefits from a tuned threshold. These results are single-seed, with per-fold standard deviations of macro F1
between 0.043 and 0.078.

\begin{table}[t]
\centering
\caption{\normalfont The three systems on the frozen native Teyit pool
(124 True, 234 False). P0 and P2-A are evaluated on 358 articles, P1 on the 357 that are both in the pool and carry a parseable verdict. $F1_T$ and $R_T$ are F1 and recall on the True
class.}
\label{tab:main}
\footnotesize
\begin{tabular}{@{}lcccc@{}}
\toprule
System & Acc & M-F1 & F1$_T$ & R$_T$ \\
\midrule
P0 \quad untrained, closed-book    & 0.690 & 0.611 & 0.437 & 0.347 \\
P1 \quad 3MFact-TR, open-book      & 0.689 & 0.651 & 0.536 & 0.516 \\
P2-A \quad fine-tuned, closed-book & \textbf{0.760} & \textbf{0.729} & \textbf{0.639} & \textbf{0.613} \\
\bottomrule
\end{tabular}
\end{table}

\subsection{Main Comparison}
Table~\ref{tab:main} shows the main result. The untrained base model P0 achieves 0.690 accuracy, slightly above the all-False rate of 0.654, but with only 0.611 macro-F1 and 0.437 F1 on True. The base model has limited knowledge beyond the majority class. Pipeline 1 equals P0 in accuracy ($b=c=44$, $p=1.00$) but adds 0.040 macro-F1 and 0.099 F1 for True. The agentic loop does not improve correctness for the majority class. The untrained model already has that. It improves the ability to recognise True claims at all.

Fine-tuning adds more. The Fine-Tuned Model in its native-only condition,
P2-A, is trained on 286 articles. It reaches 0.760 accuracy and 0.729 macro-F1, and exceeds Pipeline 1 significantly ($b=62$, $c=37$, $p=.015$). Its bootstrap interval for macro-F1 is [0.680, 0.778], against [0.597, 0.701] for Pipeline 1. It also exceeds P0 significantly ($b=50$, $c=25$, $p=.005$). The three systems place the agentic design
between an untrained model and a supervised one, closer to the untrained
end.

\subsection{Where the Fine-Tuned Advantage Comes From}
We train a claim-only variant of P2-A using the same methodology.  We blank all the template fields except the claim. The video description,  transcript, title, platform, and date all become the missing-field placeholder. We apply the same transformation to the training and the test inputs. This variant achieves 0.749 accuracy and 0.721 macro-F1. It is statistically indistinguishable from the full-input model ($b=41$, $c=37$, $p=.734$). The model mostly learns from how Turkish fact-checked claims are written. This changes how we interpret the main comparison. The gap between P2-A and Pipeline 1 measures surface signal in the claim texts as much as verification capacity.

\subsection{Ablations}

Table~\ref{tab:ablation} summarises both. We examine two internal choices in addition to the input ablation mentioned above: the SIC signal and the training mixture.  Removing SIC improves every headline metric. Accuracy rises from 0.689 to 0.709, macro-F1 from 0.651 to 0.694, and recall of True from 0.516 to 0.710. However, the paired test shows no statistically significant difference. The advisory helps on 40 articles and hurts on 47 ($p=.52$).  We see that SIC buys majority-class precision by suppressing True verdicts. Therefore, we use it only as a hint for the Reasoner rather than letting SIC make the final decision. 

Table~\ref{tab:mixture} reports the training-mixture conditions. On the data side, we found that balanced augmentation had a neutral effect. The native-only model remains nearly unchanged when the 731 translated Snopes articles (P2-B1) are added. However, indiscriminate augmentation proved highly harmful. Under an epoch-matched control (P2-A34, 2.11 epochs against P2-B1's 2.10), the native-only model retains a 22 point advantage in True recall over P2-B. The collapse of P2-B is therefore not an artefact of the step budget. When we added 792 all-False native articles and 215 all-True synthetic articles (P2-B), the model learned a false shortcut by simply linking the data source and the label. The model's True recall dropped from 0.613 to 0.306. This failure confirms exactly why balancing the dataset is so critical to prevent shortcut learning (Section~\ref{sec:dataset}).

\begin{table}[t]
\centering
\caption{\normalfont Ablations. Each block removes one component from its own
full configuration. $p$ is the exact McNemar test against that configuration.}
\label{tab:ablation}
\footnotesize
\begin{tabular}{@{}lccccc@{}}
\toprule
Condition & Acc & M-F1 & F1$_T$ & R$_T$ & $p$ \\
\midrule
P1 full                & 0.689 & 0.651 & 0.536 & 0.516 & --- \\
\quad w/o SIC advisory & 0.709 & 0.694 & 0.629 & 0.710 & 0.52 \\
\midrule
P2-A full input        & 0.760 & 0.729 & 0.639 & 0.613 & --- \\
\quad claim only       & 0.749 & 0.721 & 0.634 & 0.629 & 0.73 \\
\bottomrule
\end{tabular}
\end{table}

\begin{table}[t]
\centering
\caption{\normalfont Training mixture. All conditions share the recipe and the
evaluation pool; only the training data and the step budget differ. $p$ is the
exact McNemar test against P2-A. P2-B loses eight points of macro-F1 without
becoming distinguishable article by article, because its errors concentrate
on the True class rather than raising the total error count.}
\label{tab:mixture}
\footnotesize
\begin{tabular}{@{}lrccccc@{}}
\toprule
Condition & Train & Steps & Acc & M-F1 & R$_T$ & $p$ \\
\midrule
P2-A \quad native only     & 286   & 120 & 0.760 & 0.729 & 0.613 & --- \\
P2-B1 \quad + Snopes       & 1,017 & 120 & 0.760 & 0.726 & 0.589 & 1.00 \\
P2-B \quad + all           & 2,024 & 120 & 0.749 & 0.647 & 0.306 & 0.74 \\
\midrule
P2-A34 \quad epoch-matched & 286   & 34  & 0.721 & 0.680 & 0.524 & 0.06 \\
\bottomrule
\end{tabular}
\end{table}

\begin{figure}[!t]
\centering\scriptsize
\begin{minipage}[t]{0.485\linewidth}\vspace{0pt}
\begin{tcolorbox}[colback=white,colframe=black!45,arc=1mm,boxrule=0.5pt,
  height=7.0cm,left=2.5pt,right=2.5pt,top=2.5pt,bottom=2.5pt]
\zone{1.75cm}{%
\goldbadge{badred}{FALSE}\\[2pt]
\emph{Hindistan'da sel bölgesini timsahın bastığını gösteren videonun \textbf{güncel}
olduğu iddiası}\\
(``the video showing a crocodile in the flooded region is \textbf{current}'')}
\zone{2.35cm}{\centering
\includegraphics[width=\linewidth,height=2.3cm,keepaspectratio]{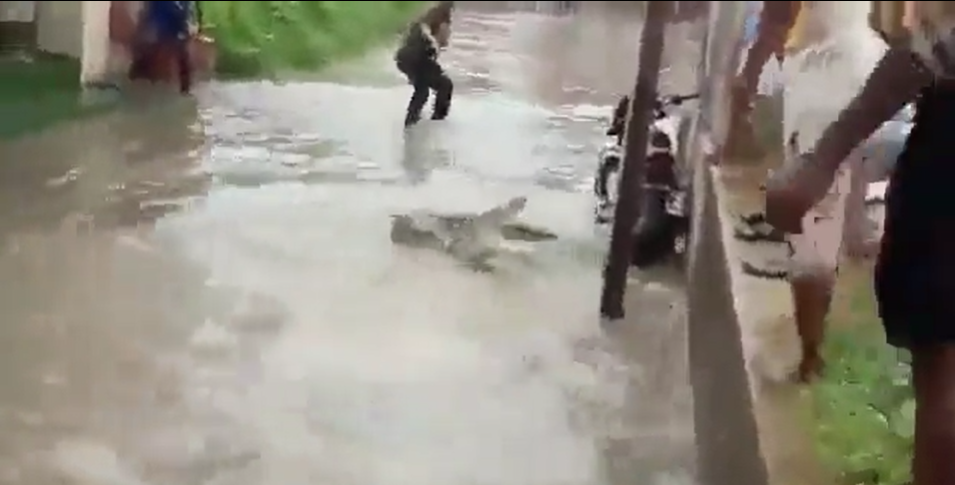}}
\begin{resbox}[height=2.40cm]
\textbf{Pipeline Result}\\[1pt]
\textbf{Verdict:} True~\textcolor{badred}{\ding{55}}\\
\textbf{Consistency:} \textsc{consistent} (0.8)\\
\textbf{Rationale:} \textcolor{badred}{``does not actively contradict the
claim''}\\
\textbf{Questions:} when, where, why\\
\textbf{Evidence:} dates the flood to 2019, unused
\end{resbox}
\end{tcolorbox}
\smallskip
{\centering (a) Silence read as permission\par}
\end{minipage}\hfill
\begin{minipage}[t]{0.485\linewidth}\vspace{0pt}
\begin{tcolorbox}[colback=white,colframe=black!45,arc=1mm,boxrule=0.5pt,
  height=7.0cm,left=2.5pt,right=2.5pt,top=2.5pt,bottom=2.5pt]
\zone{1.75cm}{%
\goldbadge{okgreen}{TRUE}\\[2pt]
\emph{Çin'de binanın içinden tren hattı geçtiği iddiası}\\
(``a railway line passes through a building in China'')}
\zone{2.35cm}{\centering
\includegraphics[width=\linewidth,height=2.3cm,keepaspectratio]{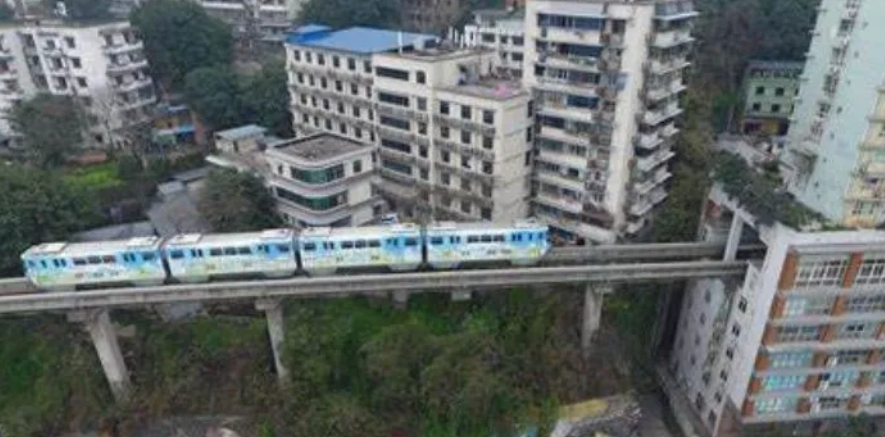}}
\begin{resbox}[height=2.40cm]
\textbf{Pipeline Result}\\[1pt]
\textbf{Verdict:} False~\textcolor{badred}{\ding{55}}\\
\textbf{Consistency:} \textsc{contradicts} (0.9)\\
\textbf{Rationale:} \textcolor{badred}{``no visual or verbal evidence of a
train''}\\
\textbf{Questions:} why\\
\textbf{Evidence:} confirms the Liziba station, unused
\end{resbox}
\end{tcolorbox}
\smallskip
{\centering (b) Silence read as contradiction\par}
\end{minipage}
 
\caption{\normalfont Two failure modes of Pipeline 1, one from each of the two largest
error classes. The description is silent on the point at issue in both cases.
In (a) the pipeline reads that silence as permission and accepts a false
claim; in (b) it reads the same silence as contradiction and rejects a true
one. Removing the consistency advisory reverses (b).}
\label{fig:errors}
\end{figure}

\subsection{Error Analysis}
\label{sec:err}
Pipeline 1 mistakes are classified into two mirror classes, as shown in Figure~\ref{fig:errors}, with one example from each. The error rate for False articles is 46.4\% when the consistency check reports \textsc{consistent} and 11.9\% when it reports \textsc{contradicts}.  61.0\% of the True articles carrying a \textsc{contradicts}
note are rejected. The direction of the advice determines which class is lost.

The description is silent on the point at issue in both cases of
Figure~\ref{fig:errors}. In (a) the recording is real and the claim lies
about time, which a visual check cannot see. The Reasoner reads the silence
as permission. In (b) the claim is true, but the recording does not show it. The Reasoner reads the same silence as contradiction. The pipeline gathered evidence and answered at least one question for all 51 False articles, showing that the issue is with evidence interpretation rather than retrieval.

\subsection{Comparison with the Original English System}
The original 3MFact reports 0.796 accuracy and 0.789 F1 on 433 English claims
using GPT-4o-mini~\citep[Table~5]{niu2025three}. Our adaptation achieves 0.689
accuracy on 357 Turkish claims. Because the two systems differed in language, dataset,
and search ecosystem, the results cannot be compared directly. Our configuration eliminates the video-QA route, which has an ablation value of about 3.5 accuracy points. Their most comparable model  is the \emph{w/o}~QA\textsubscript{LMM} row with 0.762~\citep[Table~5]{niu2025three}. We focus on the qualitative success of adapting the system, rather than a direct numeric comparison. The
multi-role design transfers to Turkish and runs end to end on open 7B
models, with no paid API and no language-specific training. The gap to the
English figures is the cost of that independence.

\section{Discussion and Limitations}
\label{sec:discussion}

This project began with a complete absence of Turkish video fact-checking resources and concludes by introducing three key contributions. The first is a 2,097 article multimodal dataset built from scratch. The second is a fully open-source verification pipeline that runs end-to-end in Turkish. The third is a controlled comparison between the two dominant approaches to the task. The most important part of this comparison is what it shows. Our results separate two questions that a single benchmark number usually merges: which system scores higher, and which score carries substance. On this pool, the fine-tuned model scores higher. However, the claim-only analysis shows where that score comes from. The model reaches almost the same result without seeing the video texts or the metadata. The label of a Turkish fact-checked claim can be predicted from how the claim is written. That signal is real and cheap to learn, but it belongs to the dataset, not to the task. Pipeline 1 does not use labels. Live evidence adds value to the True class, resulting in improvements over the untrained model. Thus, the comparison determines how much of a benchmark a model can absorb without verifying anything.

The gold labels have no part in inference. We use them only to train
Pipeline 2 and to score both systems. The true test of a design is a new claim that no fact-checker has yet answered. The two systems approach the test very differently. Pipeline 1 does not need the claim to be in any collection. It asks questions about the external world and
answers them with live search. When the evidence is not enough, it already
signals this through its sufficiency confidence and its failure records. Pipeline 2 does not have such a route. It learned the patterns of one collection, and a new claim from outside that collection gives it nothing to lean on. We met the same danger while developing TeViD. At first, every translated article was True and almost every native article was False, so a model could guess the label from the source alone. We balanced the translated portion to remove that shortcut, and the ablation confirms the danger:  the imbalanced mixture reduces True recall to 0.306. Balancing removes the source shortcut. The claim-only result shows a second shortcut, the writing style of the claims, and that one cannot be balanced away.

Several limits qualify these results. Pipeline 2 results are
single-seed, with a per-fold spread of 0.043 to 0.078. The verdict space is
binary. Real fact-checking needs at least an unverifiable class. 290 of the 358 rationales are not in Turkish, and only 56 cite the evidence
they were given. The evidence comes from the live web, so a fresh run in a different search climate may retrieve different pages. Finally, the pool holds 358 articles from a
single organisation. The absolute numbers should be read within these
bounds.

\section{Conclusion and Future Work}
\label{sec:conclusion}

We presented TeViD, the first multimodal video fact-checking dataset for
Turkish, with 2,097 fact-checked claims from three sources. On it we built two
systems that share one open base model. Pipeline 1 adapts the multi-role
3MFact framework to Turkish with three new modules and no training. Pipeline 2 fine-tunes the same model into a closed-book verdict system. The fine-tuned model scores higher, 0.729 against 0.651 macro-F1. The claim-only analysis shows that its advantage comes from how the claims are written rather than from the video.

In future work, Pipeline 1's abstention path should become a third, unverifiable verdict. Rationales should be produced in Turkish and evaluated for quality. To improve the accuracy of Pipeline 2 estimates, a multi-seed replication is recommended. Additionally, a second fact-checking source can determine how much of the surface signal survives collection changes.

\bibliography{references}
\vspace{12pt}

\newpage

\appendices

% ---------------------------------------------------------------------------
% A --- AI Statement (imzali PDF)
% ---------------------------------------------------------------------------

% ---------------------------------------------------------------------------
% B --- SIC prompt
% ---------------------------------------------------------------------------
\section{Semantic Inconsistency Checker Prompt}
\label{app:sicprompt}

The system prompt is given below in translation; the model is instructed to
write in Turkish and to return a single JSON object. Four format-only examples
are omitted for space.

\begin{quote}\footnotesize
You are a visual--claim consistency checker. You will be given a CLAIM and the
VISUAL DESCRIPTION of the video it concerns.

Turkish claims are written as ``the claim that the video shows X''. This means
the claim \emph{asserts} X. Do not read the claim in reverse.

First write: (1)~\emph{claim\_says}: what does the claim assert the video
shows, in one sentence; (2)~\emph{video\_shows}: what is actually in the video
according to the description, in one sentence.

Then answer explicitly: (3)~\emph{can\_both\_hold}: can the claim and the
video content be true at the same time? Answer yes or no. If unsure, answer
yes.

Then select one verdict:

\begin{itemize}\itemsep1pt
\item \textsc{contradicts}: the video clearly shows something else,
identifiable, in place of what is claimed --- another place, event, person or
period. The video is positively incompatible with the claim.
\item \textsc{not visible}: the claimed thing is absent from the footage, but
there is no clear opposition. The absence of something is not a contradiction.
\item \textsc{unintelligible}: the claim cannot be settled visually at all,
for example speech content, statistics, or dates.
\item \textsc{consistent}: the video supports the claim.
\end{itemize}

Rules. The \textsc{contradicts} verdict is valid only when
\emph{can\_both\_hold} is no; answering yes and then choosing
\textsc{contradicts} is inconsistent. The decisive test is whether you can say
that the claim and the video cannot both be true at the same time. If the two
can coexist, it is not a contradiction.

If a claimed detail is merely missing from the description, that is
\textsc{not visible} and not \textsc{contradicts}, because the vision model
may simply have omitted it. If you are unsure about a person's identity,
answer \textsc{not visible}, since face recognition is unreliable.

Confidence scale: a physical contradiction, where the two cannot coexist,
gives 0.9--1.0. A merely absent detail gives at most 0.4, with the verdict
\textsc{not visible}. The absence of something from a text does not make it
false.
\end{quote}

% ---------------------------------------------------------------------------
% C --- Full pipeline trace
% ---------------------------------------------------------------------------
\section{Full Pipeline Trace}
\label{app:trace}

The record below is \texttt{teyit\_ac216821165c}, the case shown in the left
panel of Fig.~\ref{fig:errors}. Turkish output is translated and long answers
are abridged.

\begin{quote}\footnotesize
\textbf{Claim.} \emph{Hindistan'da sel bölgesini timsahın bastığını gösteren
videonun güncel olduğu iddiası} (``the video showing a crocodile in the
flooded region is current''). Gold label: \textsc{False}.

\medskip
\textbf{Video information.} Headline: ``Stray crocodile captured by Agniveer
Pranin foundation during Vadodara flood''. Channel: DeshGujaratHD. Platform:
YouTube. Date: 2019-08-31.

\medskip
\textbf{1. Video description (offline).} Two people struggle with a crocodile
in the flooded garden of a house. One strikes it with a plank while the other
waits behind. The crocodile slides into the water, and a group follows it,
places a wooden board over it, and carries it to higher ground.

\medskip
\textbf{2. Consistency check.} \textsc{consistent}, confidence 0.8.
\emph{claim\_says}: the video shows a crocodile in a flooded region.
\emph{video\_shows}: two people struggling with a crocodile that is then
carried on a board. \emph{reasoning}: the crocodile is in the garden of a
house, not in a flood zone.

\medskip
\textbf{3. Question manager.} Primary dimension \emph{When}: ``When did the
Vadodara flood occur, and how does it relate to the video date?'' Reserve
dimensions: Who, What, Why, How.

\medskip
\textbf{4. Evidence and sufficiency.} The primary question is answered at 95\%
confidence: ``The Vadodara flood occurred on July 31, 2019, and the video was
uploaded on August 31, 2019, capturing the aftermath of the flood.'' The
sufficiency check is not yet satisfied, and two further rounds follow.
\emph{Where} (90\%): the rescues took place in residential areas of Vadodara
and on the campus of Maharaja Sayajirao University. \emph{Why} (90\%): the
crocodiles posed a direct threat to residents during the flood. The loop then
closes.

\medskip
\textbf{5. Reasoner.} \textsc{True}. ``The video shows a crocodile being
rescued in a garden, which is a plausible scenario during a flood. The video
does not actively contradict the claim.''

\medskip
\textbf{Outcome.} Every stage works except the last. The question manager
identifies \emph{When} as the dimension on which the claim turns, retrieval
returns an answer that dates the footage to 2019, and the sufficiency check
gathers two further rounds of confirming evidence. The Reasoner then returns
\textsc{True} on the ground that the footage does not contradict the claim,
without using the date it was given. This is the failure mode of
Fig.~\ref{fig:errors}(a) in full: the evidence needed to reject the claim was
retrieved and then not applied.
\end{quote}

\section{Worked Examples}
\label{app:worked}

\begin{figure*}[t]
\centering\footnotesize
\begin{tcolorbox}[colback=gray!12,colframe=gray!55!black,arc=1.5mm,
boxrule=0.6pt,left=3pt,right=3pt,top=2pt,bottom=2pt]
\centering\bfseries Worked examples: the two pipelines on TeViD test claims
\end{tcolorbox}
\vspace{3pt}
 
% ==================== SOL: PIPELINE 1 ====================
\begin{minipage}[t]{0.485\textwidth}\vspace{0pt}
\begin{center}\bfseries\footnotesize
Pipeline 1: 3MFact-TR (open-book)
\end{center}
\vspace{-2pt}
\begin{minipage}[t]{0.62\linewidth}\vspace{0pt}
\begin{infobox}[title=Claim]{green}
\emph{Suların çekildiğini gösteren videonun Kızıldeniz'de kaydedildiği
iddiası} (``the video of the receding waters was recorded in the Red Sea'')
\end{infobox}
\end{minipage}\hfill
\begin{minipage}[t]{0.35\linewidth}\vspace{0pt}
\begin{infobox}[title=Label]{red}
Gold: \textbf{FALSE}\\ System: \textbf{FALSE}~\checkmark
\end{infobox}
\end{minipage}
 
\vspace{3pt}
\begin{minipage}[t]{0.41\linewidth}\vspace{0pt}
\begin{infobox}[title=Video]{blue}
\centering\includegraphics[width=\linewidth,height=3.1cm,keepaspectratio]{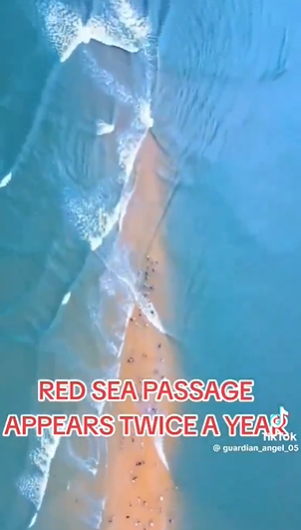}
\end{infobox}
\end{minipage}\hfill
\begin{minipage}[t]{0.56\linewidth}\vspace{0pt}
\begin{infobox}[title=Video information]{blue}
\begin{tabular}{@{}p{0.38\linewidth}p{0.56\linewidth}@{}}
\textbf{Headline} & ``Summer holiday escape on this fishbone sandbank''\\
\textbf{Channel} & New China TV\\
\textbf{Date} & 2020-07-02\\
\textbf{Platform} & YouTube\\
\textbf{Description} & places the sandbank in SE China's Fujian\\
\end{tabular}
\end{infobox}
\end{minipage}
 
\vspace{3pt}
\begin{stepbox}{\stepnum{1}~What the video shows (offline)}
(excerpt, translated) A fishbone-shaped sandbank in a sea view; as the water
recedes, wave patterns form on it and a group of people walks out to explore.
The site has drawn thousands of tourists.
\end{stepbox}
\vspace{2pt}
\begin{stepbox}{\stepnum{2}~Consistency check (SIC)}
\begin{tabular}{@{}p{0.24\linewidth}p{0.71\linewidth}@{}}
\textbf{verdict} & \textsc{consistent} (0.8); advisory only\\
\textbf{video shows} & a sea view with wave patterns\\
\textbf{reasoning} & the scenery could match the Red Sea; nothing in the
footage rules the claim out\\
\end{tabular}
\end{stepbox}
\vspace{2pt}
\begin{stepbox}{\stepnum{3}~Question manager}
\textbf{Primary (Where):} is the fishbone-shaped sandbank located in the Red
Sea, as claimed? \textbf{Reserve:} When, Why, How.
\end{stepbox}
\vspace{2pt}
\begin{stepbox}{\stepnum{4}~Sufficiency check and evidence search}
Bilingual search returns posts placing the sandbank on Dongshan Island,
Fujian, China (answer confidence 95\%). The sufficiency check stays cautious
after round one (85\%), triggers two follow-up questions, then closes.
\end{stepbox}
\vspace{2pt}
\begin{stepbox}{\stepnum{5}~Reasoner}
\textbf{Verdict:} \textsc{False}; \textbf{type:} false video description.
``The video does not actively contradict the claim [...] but the description
provides clear evidence that the location is not the Red Sea.''
\end{stepbox}
\end{minipage}
\hfill
% ==================== SAG: PIPELINE 2 ====================
\begin{minipage}[t]{0.485\textwidth}\vspace{0pt}
\begin{center}\bfseries\footnotesize
Pipeline 2: Fine-Tuned Model (closed-book)
\end{center}
\vspace{-2pt}
\begin{minipage}[t]{0.62\linewidth}\vspace{0pt}
\begin{infobox}[title=Claim]{green}
\emph{Dalila Jakupovic'in Covid-19 aşısı yüzünden maça devam edemediği
iddiası} (``Jakupovic could not continue her match because of the Covid-19
vaccine'')
\end{infobox}
\end{minipage}\hfill
\begin{minipage}[t]{0.35\linewidth}\vspace{0pt}
\begin{infobox}[title=Label]{red}
Gold: \textbf{FALSE}\\ System: \textbf{FALSE}~\checkmark
\end{infobox}
\end{minipage}
 
\vspace{3pt}
\begin{minipage}[t]{0.41\linewidth}\vspace{0pt}
\begin{infobox}[title=Video]{blue}
\centering\includegraphics[width=\linewidth,height=3.1cm,keepaspectratio]{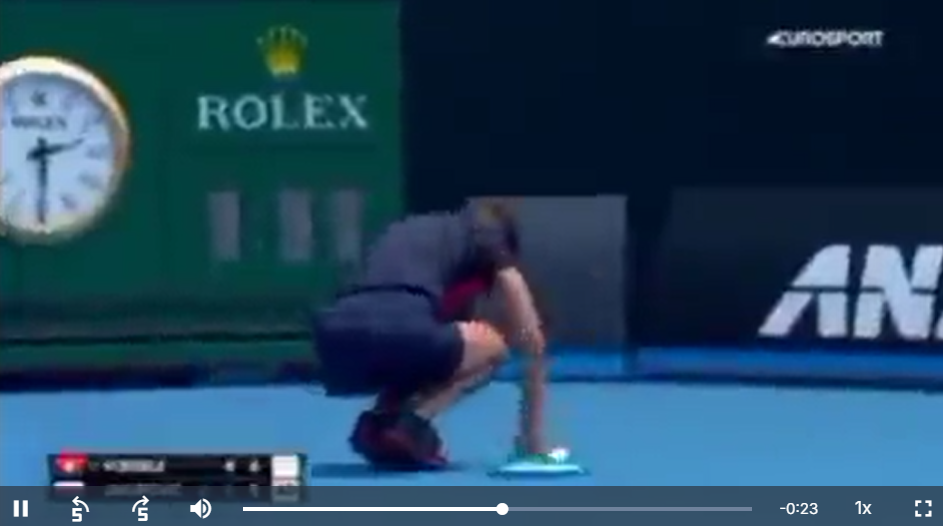}
\end{infobox}
\end{minipage}\hfill
\begin{minipage}[t]{0.56\linewidth}\vspace{0pt}
\begin{infobox}[title=Video information]{blue}
\begin{tabular}{@{}p{0.32\linewidth}p{0.62\linewidth}@{}}
\textbf{Headline} & ``Dalila Jakupovic on her collapse at Australian Open''\\
\textbf{Date} & \textbf{2020-01-15}\\
\textbf{Platform} & YouTube\\
\end{tabular}
\end{infobox}
\end{minipage}
 
\vspace{3pt}
\begin{stepbox}{\stepnum{1}~Fixed-slot input (abridged)}
\texttt{\scriptsize Iddia, Video basligi, Platform, Video tarihi:} as above\\
\texttt{\scriptsize Video aciklamasi:} ``...she suffered a coughing fit
because of the poor quality of air in Melbourne. Australia is dealing with
its worst bushfires in decades...''\\
\texttt{\scriptsize Video transkripti:} ``...this smoke is something
different...''\\
\texttt{\scriptsize Video icerigi (betimleme):} a woman interviewed in front
of Laver Arena\\
\texttt{\scriptsize Karar (True/False):}
\end{stepbox}
\vspace{2pt}
\begin{stepbox}{\stepnum{2}~Scoring (single forward pass)}
$s=\log P(\text{True})-\log P(\text{False})=-1.75$; the verdict is read from
the first answer token and no text is generated. The untrained base model
(P0), scored identically, reaches the same verdict far more confidently
($s=-15.6$).
\end{stepbox}
\vspace{2pt}
\begin{stepbox}{\stepnum{3}~Verdict}
\textbf{False} (gold: FALSE) \checkmark\; No rationale is produced. The
headline confirms the collapse but is silent on its cause; the video's own
account attributes it to bushfire smoke, and the date precedes Covid-19
vaccines.
\end{stepbox}
 
\vspace{5pt}
\begin{tcolorbox}[colback=gray!7,colframe=gray!45!black,arc=1.5mm,
boxrule=0.5pt,left=4pt,right=4pt,top=3pt,bottom=3pt]
\footnotesize\textbf{Contrast.} The left claim could only be settled by
external evidence: the footage is authentic and visually consistent, and the
lie is about location. The right claim is settled without any retrieval, from
the video's own textual account and the model's timeline knowledge. Neither
pipeline looks at the pixels once the two texts have been produced.
\end{tcolorbox}
\end{minipage}
 
\caption{\normalfont Worked examples of the two pipelines on TeViD test claims, with the
step numbers of Fig.~\ref{fig:arch}. \emph{Left:} authentic footage attached
to a false location. Visual consistency cannot detect a location lie, so the
consistency check correctly reports \textsc{consistent} and the verdict is
carried by retrieved evidence together with the video's own metadata.
\emph{Right:} the Fine-Tuned Model reconciles the claimed cause with the
video's textual account and its own timeline knowledge; no retrieval is
performed and no rationale is produced, and the frame is shown for
illustration only. Excerpts are translated from Turkish and abridged; a complete trace, for the case in Fig.~\ref{fig:errors}(a), is given in Appendix~\ref{app:trace}.}
\label{fig:worked}
\end{figure*}

\end{document}